\documentclass{article} 
\usepackage{iclr2020_conference,times}

\usepackage{amsmath,amsfonts,bm}

\def\eqref#1{equation~\ref{#1}}

\def\1{\bm{1}}

\DeclareMathAlphabet{\mathsfit}{\encodingdefault}{\sfdefault}{m}{sl}
\SetMathAlphabet{\mathsfit}{bold}{\encodingdefault}{\sfdefault}{bx}{n}

\usepackage{hyperref}
\usepackage{url}
\usepackage{subfigure}
\usepackage{graphicx}
\usepackage{comment}
\usepackage{todo}

\title{Evaluation of Model Selection for Kernel Fragment Recognition in Corn Silage}

\author{Christoffer B\o gelund Rasmussen \& Thomas B. Moeslund  \\
Department of Architecture, Design \& Media Technology\\
Aalborg University\\
Rendsburggade 14, 9000 Aalborg, Denmark\\
\texttt{\{cbra,tbm\}@create.aau.dk} \\
}

\iclrfinalcopy 
\begin{document}

\maketitle

\begin{abstract}
Model selection when designing deep learning systems for specific use-cases can be a challenging task as many options exist and it can be difficult to know the trade-off between them. Therefore, we investigate a number of state of the art CNN models for the task of measuring kernel fragmentation in harvested corn silage. The models are evaluated across a number of feature extractors and image sizes in order to determine optimal model design choices based upon the trade-off between model complexity, accuracy and speed. We show that accuracy improvements can be made with more complex meta-architectures and speed can be optimised by decreasing the image size with only slight losses in accuracy. Additionally, we show improvements in Average Precision at an Intersection over Union of 0.5 of up to 20 percentage points while also decreasing inference time in comparison to previously published work. This result for better model selection enables opportunities for creating systems that can aid farmers in improving their silage quality while harvesting.
\end{abstract}

\section{Introduction}
Computer vision systems for quality inspection are widespread throughout agriculture and many other industries. Deep learning has become the driving force in many applications largely due to advantages such as potentially high accuracy and ease of use due to the large number of open source libraries. The common methodology for training the networks is either to adapt an open-source network or for an author to design their own network. However, it can be difficult to choose which network is best for a specific task as it often comes with a trade-off between complexity, accuracy and speed. Therefore, in this work our contribution is showing a systematic approach is create an overview over the trade-off for a specific agricultural task of corn kernel fragment recognition from corn silage harvested from a forage harvester. In corn silage kernels must be cracked sufficiently such that when used as fodder for dairy cows the starch content is easily ingested and milk yield can be optimised \citep{johnson03}. An recognition system for high quality can help farmers use their machine optimally, avoiding both quality decreasing by up to 25\% and inefficient usage of diesel fuel \citep{marsh13}. Furthermore, such systems can help solve the potential food crisis as the population is expected to reach 9.1 billion in 2050 \citep{un09}.

This work extends upon that done in \cite{Rasmussen19} where it was shown that kernel fragment shape and size characteristics could be measured with Convolutional Neural Networks (CNNs) for bounding-box detection and instance segmentation, however, only a single form of each was trained and it is unknown if these architectures are optimal. In \cite{huang17} the trade-off between speed and accuracy was explored for CNN-based object detectors. Whilst comprehensive and useful as the open-source implementations are available through TensorFlow object detection API, networks are trained and evaluated on the large COCO benchmark dataset \citep{mscoco} and it is not as clear what the trade-off is for a specific use-case on a smaller scale like kernel fragmentation. We provide an overview of the trade-off for the kernel recognition by training variants of three meta-architectures of increasing complexity with the API from \cite{huang17} and explore different feature extractors and input image resolutions. This allows us to show an approach to determine optimal model design choices for CNN-based kernel fragment recognition.

\section{DATA}
\label{sec:data}
The data used to train and test the networks are the same as that used in \cite{Rasmussen19} and consist of RGB images of silage taken post-harvest. Typically, kernel processing evaluation requires the separation of kernels and stover (leaves and stalks) either through manual means as in \citep{csps, psps} followed by sieving measurements or sieving estimation with image processing \citep{silagesnap}. However, the manual separation step can be cumbersome making it problematic for a farmer whilst harvesting. Therefore, in \cite{Rasmussen19} images and annotations were collected of non-separated corn silage for a direct measurement. 

The dataset consists of a total of 2043 images with 11601 kernel fragment annotations. A notable difference in this work compared to \cite{Rasmussen19} is a validation set is added to combat overfitting whilst training by evaluating a model variant with the lowest validation loss. In \cite{Rasmussen19} the data was split 60\% for training and 40\% for testing, here we keep the same training set but evenly split the original test set such that validation and test cover 20\% each. For the variation of image sizes when training and testing models images are resized from the original images dimensions of 640$\times$1280 to either 600$\times$1200, 400$\times$730 or 200$\times$365 using bilinear interpolation.

\section{CNN META-ARCHITECTURES}
The TensorFlow object detection API provides a number of options for meta-architectures and includes pre-trained models with different backbone feature extractors and hyperparameters. Hyperparameters for the training of our models remained unchanged to the configurations files provided in the API, apart from the learning rate being decreased by a factor of 10 as only fine-tuning is performed. Networks are trained using TensorFlow 1.13.1 on an machine containing an NVIDIA Titan XP and GTX 1080Ti.

The first meta-architecture adopted is the Single Shot Multibox Detector (SSD) and is an efficient single-stage bounding-box detector. SSD has a competitive accuracy whilst running much faster than other more complex networks. For the varying complexity of feature extraction within SSD we adopt MobileNetv1 \citep{mobilenetv1}, MobileNetv2 \citep{mobilenetv2} and InceptionV2 \citep{inceptionv2}. Next, we train Faster R-CNN, a two-stage bounding-box detector that utilises the Region Proposal Network (RPN) to produce candidate proposals whose boxes are regressed and classified. For Faster R-CNN we train variants with Inceptionv2, ResNet50 and ResNet101 from \cite{resnet}. Lastly and most complex is the instance segmentation network Mask R-CNN \citep{maskrcnn}. The network is an extension of Faster R-CNN but with the added ability of producing masks for prediction. As the RPN is also part of Mask R-CNN the network is also able to output bounding-boxes, thus both forms will be evaluated. The feature extractors trained for Mask R-CNN are also Inceptionv2, ResNet50 and ResNet101. 

\section{RESULTS}
\label{sec:results}

The results in Table \ref{tab:results} are based upon a subset of the COCO metrics where the models with bounding-box predictions can be seen in first section and segmentation models in the second section. Additionally, we show the AP@0.5 results from \cite{Rasmussen19} for R-FCN \cite{rfcn} with ResNet101 and the MNC \citep{mnc} with AlexNet \citep{alexnet}. As mentioned in Section \ref{sec:data}, we altered the test set such that a validation set is also available. Therefore, the results are not calculated on the exact same images as in \cite{Rasmussen19} but we argue that the new test set is large enough such that the results are comparable.

\begin{table}[h]
\scriptsize
	\caption{Results of the models on the test set. The bounding-box outputs are evaluated are shown in the first section followed by the segmentation outputs.}
	\label{tab:results}
	\begin{center}
		\begin{tabular}{llllll}
			\multicolumn{1}{c}{\bf MODEL}  &\multicolumn{1}{c}{\bf IMAGE SIZE} &		\multicolumn{1}{c}{\bf AP} &		\multicolumn{1}{c}{\bf AP@0.5} &	\multicolumn{1}{c}{\bf AR@100} &		\begin{tabular}[c]{@{}l@{}}\textbf{INFERENCE}\\ \textbf{TIME (ms)}\end{tabular}   
			\\ \hline \\
			R-FCN ResNet101 \citep{Rasmussen19}     	 & 600$\times$1200 & NA & 34.0 & NA & 101.0 \\
			SSD MobileNetV1         	 & 600$\times$1200 & 20.3 & 43.5 & 41.8 & 18.8  \\
						            	 & 400$\times$730  & 16.0 & 34.9 & 38.9 & 13.8  \\
							       	     & 200$\times$365  & 9.7 & 27.9 & 30.8 & \textbf{13.2}  \\
			SSD MobileNetV2         	 & 600$\times$1200 & 22.2 & 47.0 & 44.5 & 21.1  \\
						            	 & 400$\times$730  & 22.1 & 48.7 & 43.7 & 15.6  \\
							       	     & 200$\times$365  & 13.5 & 35.9 & 33.9 & 15.4  \\				       	     
			SSD InceptionV2        	 	 & 600$\times$1200 & 19.3 & 41.3 & 39.1 & 24.8  \\
						           	 	 & 400$\times$730  & 19.6 & 46.3 & 37.9 & 19.6  \\
							       	 	 & 200$\times$365  & 14.6 & 36.9 & 32.4 & 18.3  \\
			Faster R-CNN InceptionV2	 & 600$\times$1200 & 25.6 & 51.9 & 45.1 & 51.1  \\
						            	 & 400$\times$730  & 24.5 & 52.5 & 41.5 & 44.1  \\
							        	 & 200$\times$365  & 15.7 & 39.4 & 27.5 & 41.6  \\
			Faster R-CNN ResNet50		 & 600$\times$1200 & 24.5 & 51.1 & 45.8 & 96.8  \\
						           	     & 400$\times$730  & 20.5 & 44.5 & 38.0 & 84.8  \\
							         	 & 200$\times$365  & 10.7 & 29.2 & 23.3 & 76.2  \\
			Faster R-CNN ResNet101	 	 & 600$\times$1200 & 25.5 & 52.1 & 45.3 & 112.4  \\
						            	 & 400$\times$730  & 22.0 & 47.1 & 40.6 & 92.4  \\
							         	 & 200$\times$365  & 11.1 & 28.5 & 22.9 & 81.9  \\
			Mask R-CNN InceptionV2	 	 & 600$\times$1200 & 26.0 & 52.7 & 46.5 & 129.8  \\
						             	 & 400$\times$730  & 24.6 & 50.7 & 43.5 & 94.5  \\
							        	 & 200$\times$365  & 16.4 & 39.0 & 29.4 & 68.5  \\
			Mask R-CNN ResNet50			 & 600$\times$1200 & 26.4 & 50.7 & 49.2 & 316.6  \\
						            	 & 400$\times$730  & 26.4 & 51.2 & 46.5 & 256.8  \\
							        	 & 200$\times$365  & 13.4 & 30.0 & 27.0 & 214.7  \\
			Mask R-CNN ResNet101		 & 600$\times$1200 & 26.9 & 52.4 & \textbf{50.1} & 381.5   \\
						            	 & 400$\times$730  & \textbf{27.5} & \textbf{54.0} & 47.8 & 281.1  \\
							        	 & 200$\times$365  & 16.0 & 35.6 & 34.5 & 222.0  \\
							        	 			\\ \hline \\
			MNC AlexNet \citep{Rasmussen19}              	 & 600$\times$1200 & NA & 36.1 & NA & 87.0 \\
			Mask R-CNN InceptionV2	 	 & 600$\times$1200 & 23.3 & 51.5 & 41.2 & 129.8  \\
						             	 & 400$\times$730  & 21.7 & 49.6 & 38.2 & 94.5  \\
							        	 & 200$\times$365  & 14.1 & 36.7 & 24.8 & \textbf{68.5}  \\
			Mask R-CNN ResNet50			 & 600$\times$1200 & 23.7 & 49.8 & 43.6 & 316.6  \\
						            	 & 400$\times$730  & 24.2 & 50.7 & 42.0 & 256.8  \\
							        	 & 200$\times$365  & 12.2 & 30.1 & 24.8 & 214.7  \\
			Mask R-CNN ResNet101		 & 600$\times$1200 & 25.3 & 52.0 & \textbf{46.4} & 381.5  \\
						            	 & 400$\times$730  & \textbf{26.1} & \textbf{53.8} & 44.4 & 281.1  \\
							        	 & 200$\times$365  & 14.8 & 35.6 & 28.7 & 222.0  \\

        \end{tabular}
    \end{center}
\end{table}

\begin{figure}[h]
    \begin{center}
        \includegraphics[width=0.92\textwidth]{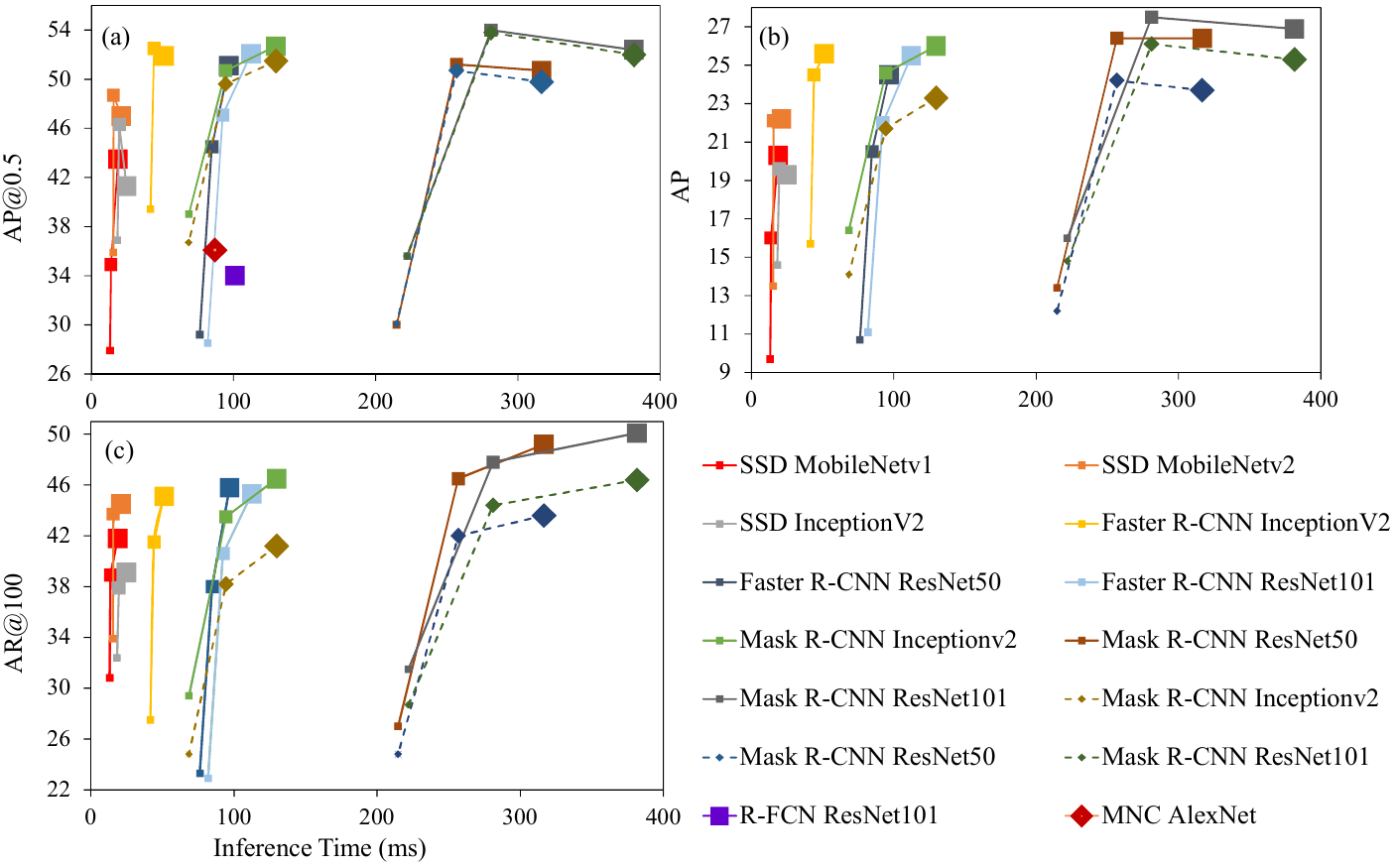}
            \caption{Results from the model variants for AP@0.5 (a), AP (b) and AR@100 (c) against inference time on an NVIDIA Titan XP. Models producing bounding-box outputs are shown with a solid line and square points and segmentation outputs are shown with a dashed line and diamond points. The increase in image size is shown by an increase in the size of the respective points.}
            \label{fig:metricsplot}
    \end{center}
\end{figure}

The results in Table \ref{tab:results} are visualised in Figure \ref{fig:metricsplot} where we show the AP@0.5 in (a), AP in (b) and AR@100 in (c) all against the inference time of the models. Firstly, we see a significant improvement in the AP@0.5 in comparison to the R-FCN model from \cite{Rasmussen19} in addition to a decrease in inference time for all SSD variants and some of the Faster R-CNNs and Mask R-CNNs. The models trained in this work have an AP@0.5 of around 20 percentage points higher, while running at up to 5-8$\times$ faster for bounding-boxes. However, the segmentation variants proved to be slower than previous with only the Mask R-CNN Inceptionv2 at image size 200$\times$365 running 1.27$\times$ faster and improving AP@0.5 by 0.6 percentage points in comparison to the MNC model from \cite{Rasmussen19}. However, improvements of up to 17.7 percentage points are seen for more complex models but at a cost of increased inference time.

Comparing the varying meta-architecture complexity we see that there is a slight gain in the metrics when evaluating bounding-box outputs. However, this comes at a cost of inference time, especially between Faster R-CNN and Mask R-CNN. Within each meta-architecture we see slight differences between feature extractors. At 600$\times$1200 AP for SSD improves by 9.4\% from MobileNetv1 to MobileNetv2 but falls for Inceptionv2, Faster R-CNN increases by 4.5\% from Inceptionv2 to ResNet101 and Mask R-CNN by 3.5\% from Inceptionv2 to ResNet101. This shows that less is gained spending time on determining the optimal architecture for feature extraction in comparison to choosing the meta-architecture. This is in contrast to the findings in \cite{huang17} where large improvements could be made, for example, Faster R-CNN had a 70\% increase in AP on the MS COCO test set over the evaluated feature extractors. Finally, we do see improvements in the metrics when increasing the image size from 200$\times$365 to 400$\times$730, but not as much when between 400$\times$730 and 600$\times$1200. Additionally, a significant increase in inference time is seen for most meta-architectures when the image size is at the largest. 

\begin{figure}[h]
    \begin{center}
        \includegraphics[width=0.92\textwidth]{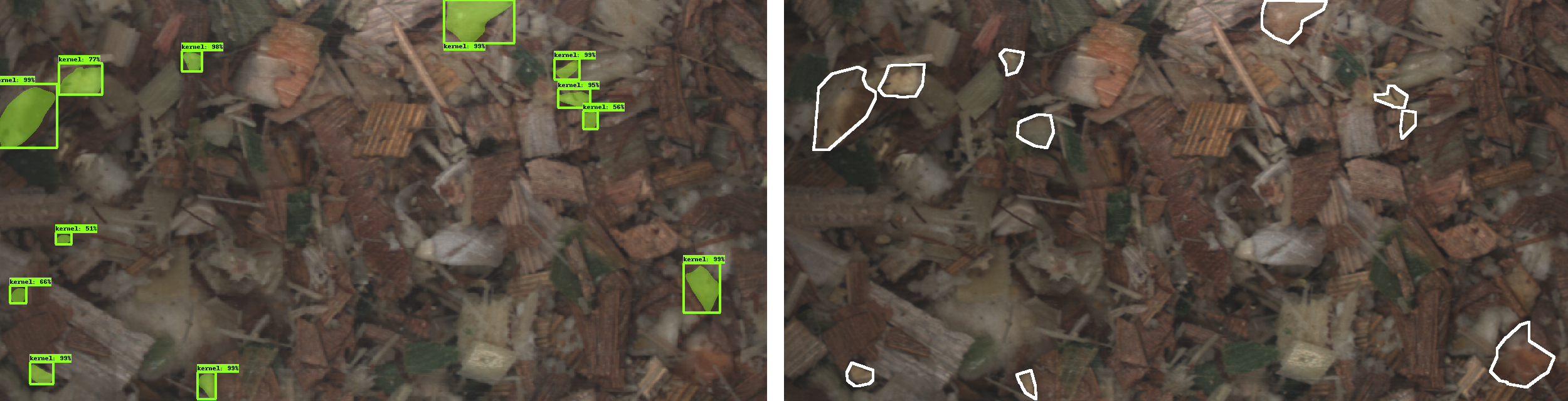}
            \caption{Left: Mask R-CNN ResNet101 (400x730) predictions. Right: Ground truth annotations.}
            \label{fig:det_ex}
    \end{center}
\end{figure}

Lastly, an example image with predictions from the best performing model with respect to AP and AP@0.5 can be seen in Figure \ref{fig:det_ex}.

\section{CONCLUSIONS}
In this work we have shown a systematic approach to train object recognition networks towards the task of kernel fragment recognition in corn silage whilst providing an overview of the trade-off in complexity, accuracy and speed. We show that slight improvements in AP and AR can be made by adopting more complex meta-architectures but at a larger cost of inference time. For all models the gain in AP and AR from a small to a medium image size was considerable, however, was minimal or worse when increasing onwards to a larger size. Minimal improvements could be made when altering the feature extractor for each meta-architecture, a contrast to findings on COCO in \citep{huang17} We propose that this approach can be transferred to other similar domains where training data can be sparse in order select an appropriate model and speculate that these design choices for our models could be directly transferred to tasks with similarities in images, such as high amounts of clutter and occlusion. The improvements in kernel fragment recognition through better model selection open possibilities for a more efficient and robust system for farmers to obtain improved yields.


\subsubsection*{Acknowledgements}
This work was funded by Innovation Fund Denmark under Grant 7038-00170B.

\bibliography{iclr2020_conference}
\bibliographystyle{iclr2020_conference}


\end{document}